\documentclass[pmlr]{jmlr}% new name PMLR (Proceedings of Machine Learning)

\usepackage{longtable}% for long tables, and
\usepackage{subcaption}
\usepackage[croatian]{babel}

\usepackage{booktabs}
\usepackage[load-configurations=version-1]{siunitx} % newer version
\theorembodyfont{\upshape}
\theoremheaderfont{\scshape}
\theorempostheader{:}
\theoremsep{\newline}

\jmlrvolume{1}
\jmlryear{2019}
\jmlrworkshop{NeurIPS2019 Disentanglement Challenge}

\title[Inadequacy of Disentanglement Metrics]{A Preliminary Study of Disentanglement  With Insights on the Inadequacy of Metrics}

  \author{\Name{Amir H. Abdi} \Email{amirabdi@ece.ubc.ca}\\
   \Name{Purang Abolmaesumi} \Email{purang@ece.ubc.ca}\\
   \Name{Sidney Fels} \Email{ssfels@ece.ubc.ca}\\
   \addr Electrical and Computer Engineering Department, University of British Columbia}

\begin{document}

\maketitle

\begin{abstract}
% A disentangled representation captures the independent features of the data.
% in such a way that if one feature changes, the others remain unaffected.
Disentangled encoding is an important step towards a better representation learning. 
However, despite the numerous efforts, there still is no clear winner that captures the independent features of the data in an unsupervised fashion.
In this work we empirically evaluate the performance of six 
unsupervised disentanglement approaches on the \texttt{mpi3d\_toy} dataset curated and released for the NeurIPS 2019 Disentanglement Challenge.
The methods investigated in this work are
$\beta$-VAE, Factor-VAE, DIP-I-VAE, DIP-II-VAE, Info-VAE, and $\beta$-TCVAE. 
The capacities of all models were progressively increased throughout the training and the hyper-parameters were kept intact across experiments.
The methods were evaluated based on five disentanglement metrics, namely, DCI, Factor-VAE, IRS, MIG, and SAP-Score.
Within the limitations of this study, the $\beta$-TCVAE approach was found to outperform its alternatives with respect to the normalized sum of metrics.
However, a qualitative study of the encoded latents reveal that there is not a consistent correlation between the reported metrics and the disentanglement potential of the model.

\end{abstract}

\section{Introduction}
\label{sec:intro}

Unsupervised disentanglement is an open problem in the realm of representation learning, incentivized around interpretability~\citep{Lake2016,Bengio2013}. 
% A disentangled representation is expected to contain all the information of the data in an abstract fashion~\citep{Lake2016,Bengio2013}.
% The majority of works interpret  disentanglement as a statistically independent factorial representation~\citep{betatcvae}
%  one feature of the data changes, only one factor of the representation is affected.
% such approach is shown not to be generalizable~\citep{BETAVAE2019}.
A disentangled representation is a powerful tool in transfer learning, few shot learning, reinforcement learning, and semi-supervised learning of downstream tasks~\citep{GoogleAssumptions,PetersJanzingSchoelkopf17,Bengio2013}.

Here, we compare the disentanglement performance of variational methods. 
% Here, we investigate the performance of some of the promising  disentanglement methods from the family of variational autoencoders (VAE).
The methods are evaluated based on five relatively established disentanglement metrics on the simplistic rendered images of the \texttt{mpi3d\_toy} dataset curated and released for the NeurIPS 2019 Disentanglement Challenge~\citep{gondal2019transfer}.

\section{Methods}

\subsection{Pre-training}

To mitigate the sensitivity of the models  to the initial state, as suggested by the findings of \citet{GoogleAssumptions},
an autoencoder model was pre-trained with the conventional VAE objective~\citep{VAE} on the \texttt{mpi3d\_toy} dataset.
% The learned parameters of the converged pre-trained model was then used as the initial state of 
This approach guaranteed that models did not collapse  into a local minima with little to no reconstruction.
% It also mitigated the sensitivity of the final model to the initial state as suggested by the findings of \citet{GoogleAssumptions}, and 
It also facilitated the training process given the constraints on the length of training by the challenge.

\subsection{Objective Function}
In this preliminary study, we implemented the variational objective functions proposed by the following methods: $\beta$-VAE~\citep{BetaVAE}, $\beta$-TCVAE~\citep{betatcvae}, Factor-VAE~\citep{FactorVAE}, Info-VAE~\citep{InfoVAE}, DIP-I-VAE,  and DIP-II-VAE~\citep{DIPVAE}.

% Among these objectives, the  $\beta$-TCVAE formulation was found to achieve the best performance across multiple metrics. Therefore, in this section, the details of a simplified version of it used in our implementation will be discussed.

In $\beta$-TCVAE, the mutual information between the data variables and latent variables are maximized, while the mutual information between the latent variables are minimized. 
Defining $x_n$ as the $n$th sample of the dataset, 
the evidence lower bound (ELBO) of this objective can be simplified as follows\footnote{The $\alpha$ and $\gamma$ hyper-parameters of the original formulation are assumed to be 1.}
\begin{equation}
\label{eq:obj}
\mathcal{L}_{\beta\text{-TCVAE}} = \mathbb{E}_{q}[\text{log}~p(x_n|z)] -  
\mathcal{D}(q(z|x_n), p(z)) - 
(\beta - 1) KL(q(z)||\prod_j q(z_j))~,
% _{\beta\text{-TCVAE}}
\end{equation}
where $z_j$ denotes the $j$th dimension of the latents.
In the above equation, the first term is the reconstruction loss. The second term is the distance between the assumed prior distribution of the latent space and the empirical posterior latent distribution. 
The last term is an indication of the total correlation (TC) between the latent variables which is a generalization of the mutual information for more than two variables~\citep{TC1960}.

\subsection{Progressive Capacity Increase}

A total capacity constraint which limits the KL divergence between the posterior latent distribution and the factorized prior can encourage the latent representation to be more factorised.
However, this will act as an information bottleneck for the reconstruction task and  results in a blurry reconstruction.
Thus, progressively increasing the information capacity of VAE during training can help facilitate the robust learning of the factorized latents~\citep{Capacity}. 
This is achieved by introducing the capacity term $C$ and defining the distance between distributions as the absolute deviation from $C$:
\begin{equation}
\mathcal{D}(q(z|x_n), p(z)) =    | KL (q(z|x_n)||p(z)) - C| ~.
\end{equation}
Gradually increasing $C$ has an annealing effect on the constraint and increases the reconstruction capacity of the model.

\section{Experiments and Results}

% We compared the $\beta$-TCVAE learning algorithm against the following well-investigated disentanglement approaches in the VAE family: $\beta$-VAE, Factor-VAE, Info-VAE, DIP-I-VAE,  and DIP-II-VAE.
For each learning algorithm, the hyper-parameter sub-spaces were independently searched.
However, in order for the results reported here to be comparable,
the hyper-parameters were kept intact in between the following experiments.

The input images were $64\times64$ pixels and the latent space was of size 20.
The model capacity parameter, $C$, was initiated at zero and gradually increased up to 25 over 2000 iterations. 
Learning rate was initiated at 0.001 and was reduced by a factor of 0.95 when the loss function (\equationref{eq:obj}) did not decrease after two consecutive epochs, down to a minimum of 0.0001 . 
Batch size was set to 64.
Optimization was carried out using the Adam optimizer with the default parameters $\beta1=0.9$ and $\beta2=0.999$.
The network architectures and other hyper-parameters are detailed in Appendix \ref{apd:hyper-parameters}.

The trained models were evaluated based on five evaluation metrics, namely, DCI, FactorVAE metric, IRS, MIG, and SAP-Score.
Results of these evaluations are presented in Table~\ref{tab:results}.
The non-ignored latent variables of each method are traversed and the results are visualized in Appendix \ref{apd:traverse}.
Moreover, the evaluation logs during model training are visualized in Appendix \ref{apd:training-log}.

All the models and experiments were implemented using the PyTorch deep learning library and packaged under the Disentanglement-PyTorch repository \url{https://github.com/amir-abdi/disentanglement-pytorch}.

\begin{table}[hbtp]
\floatconts
  {tab:results}
  {\caption{Disentanglement methods evaluated based on DCI, SAP, FactorVAE, MIG and IRS. Normalized Sum: Due to the inconsistencies in the scale of different metrics, each value is normalized based on the maximum of their column and summed for each method. }}
  {\begin{tabular}{lcccccc}
  \toprule
  \bfseries Method & \bfseries DCI & \bfseries FactorVAE  & \bfseries  SAP & \bfseries MIG & \bfseries IRS & \bfseries Normalized Sum\\
  \midrule
  $\beta$-TCVAE & \bfseries 0.392 &	0.458	& 0.132 & 	0.203 & \bfseries 0.646 & \bfseries 4.706
\\
  Factor-VAE & 0.389 & 0.449 & \bfseries 0.136 &  0.203 & 0.577 & 4.611 \\
  $\beta$-VAE & 0.373 &	0.501	&0.135 & \bfseries	0.212 &	0.517 & 4.599 \\
  Info-VAE & 0.381 &	0.523 &	0.128 &	 0.210 &	0.514 & 4.591 \\
  DIP-VAE-I & 0.385 &	\bfseries 0.587	& 0.127 &	0.188 &	0.358 & 4.351 \\
  DIP-VAE-II & 0.359 & 	0.584 &	0.111 &	0.163 &	0.340 & 4.023 \\
  \bottomrule
  \end{tabular}}
\end{table}

\section{Discussion}

In this work we compared the degree of disentanglement in latent encodings of six variational learning algorithms, namely, 
$\beta$-VAE, Factor-VAE, DIP-I-VAE, DIP-II-VAE, Info-VAE, and $\beta$-TCVAE. 
The empirical results (Table~\ref{tab:results}) point to $\beta$-TCVAE being marginally the superior option and, consequently, chosen as the best performing approach.
However, a qualitative study of the traversed latent spaces (Appendix~\ref{apd:traverse}) reveals that none of the models encoded a true disentangled representation.
Lastly, although the DIP-VAE-II model is under performing according to the quantitative results, it has the least number of ignored latent variables with a  promising latent traversal compared to other higher performing methods (Appendix~\ref{apd:traverse}).
% Therefore, it is possible that a particular combination of parameters happens to achieve superior results.
As a result of these inconsistencies, we find the  five   metrics utilized in this study inadequate for the purpose of disentanglement evaluation.

Among the limitations of this study is the insufficient search of the hyper-parameters space for all the six learning algorithms. 
Moreover, the NeurIPS 2019 Disentanglement Challenge imposed an 8-hour limit on the training time of the models which we found to be insufficient.
% Such constraint 
% narrowed the training algorithm's ability  to explore the model space, which in turn 
% limited the generalizability of the  unsupervised learning algorithm which is expected to perform on any given dataset.
% We found this limit constraining to the model training. 
This, while the maximum number of iterations was set to 200k in our experiments, this value was limited to 100k in the submissions made to the challenge portal.

\bibliography{jmlr-sample}

\pagebreak
\appendix

\section{Model Details}\label{apd:hyper-parameters}
~

\subsection{Architectures of the Neural Networks}
The encoder neural network in all experiments consisted of 5 convolutional layers with strides of 2, kernel sizes of $3\times3$, and number of kernels gradually increasing from 32 to 256. 
The encoder ended with a dense linear layer which estimated the posterior latent distribution as a parametric Gaussian.
The decoder network consisted of one convolutional followed with 6 deconvolutional (transposed convolutional) layers, with kernel sizes of 4, strides of 2, and the number of kernels gradually decreasing from 256 down to the number of channels of the image space. 
ReLU activations were used throughout the architecture, except for the last layers of the encoder and decoder networks. 

\subsection{Hyper-parameters}

\begin{table}[h]
\floatconts
  {tab:specs}
  {\caption{The hyper-parameters used to train each disentanglement method including the method-specific parameters and those shared among all models.}}
  {\begin{tabular}{ll}
  \toprule
  \bfseries Method & \bfseries Parameters\\
  \midrule
  $\beta$-TCVAE & $\beta=2.0$\\
  $\beta$-VAE &  $\beta=2.0$\\
  Info-VAE & $\lambda=1000$\\
  DIP-I-VAE & $\lambda_d=10,~ \lambda_od=1.0$\\
  DIP-II-VAE & $\lambda_d=10,~ \lambda_od=1.0$\\
  Factor-VAE & $\gamma=2.0$\\
\midrule
  Shared & Batch Size=64, \\
  & LR=$0.001 \rightarrow 0.0001$ by a factor of $0.95$, \\
  & C=$0 \rightarrow 25$ over 2000 steps, \\
  & Adam$_{\beta1}=0.9$, Adam$_{\beta2}=0.999$\\
  & Latent Size=20, Image Size=$64\times64$ \\
  \bottomrule  
  \end{tabular}}
\end{table}

\pagebreak
\section{Traversed Latent Space of Trained Models}\label{apd:traverse}

\begin{figure}[htbp]
\floatconts
  {fig:info-vae}
  {\caption{Traversed non-ignored latents of the trained \textbf{Info-VAE} model on a random sample of the \texttt{mpi3d\_toy} dataset.}}
  {\includegraphics[width=1.0\linewidth]{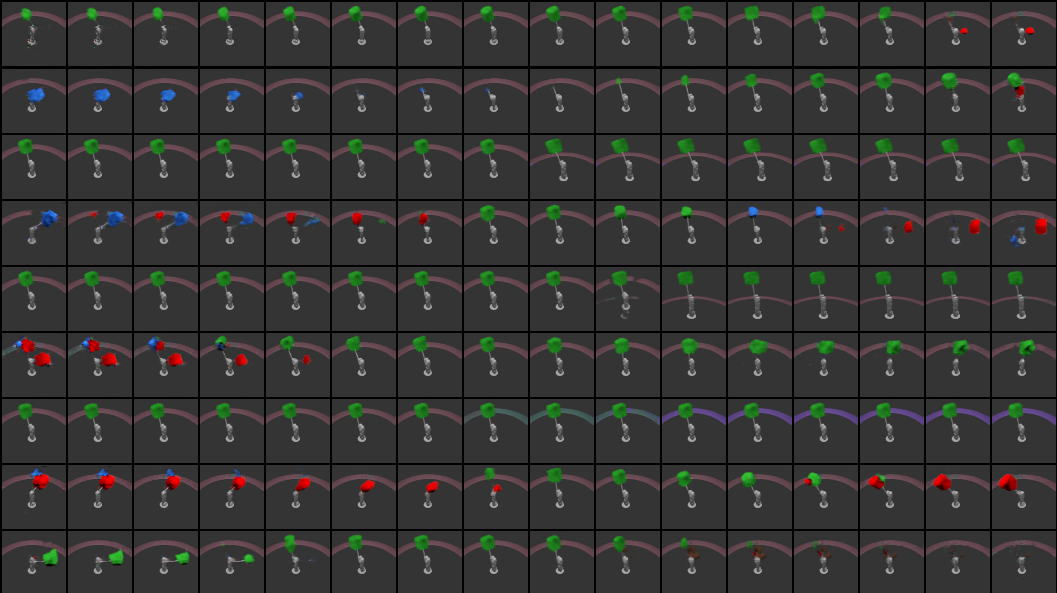}}
\end{figure}

\begin{figure}[!ht]
\floatconts
  {fig:beta-tcvae}
  {\caption{Traversed non-ignored latents of the trained  \textbf{$\beta$-TCVAE} model on a random sample of the \texttt{mpi3d\_toy} dataset.}}
  {\includegraphics[width=1.0\linewidth]{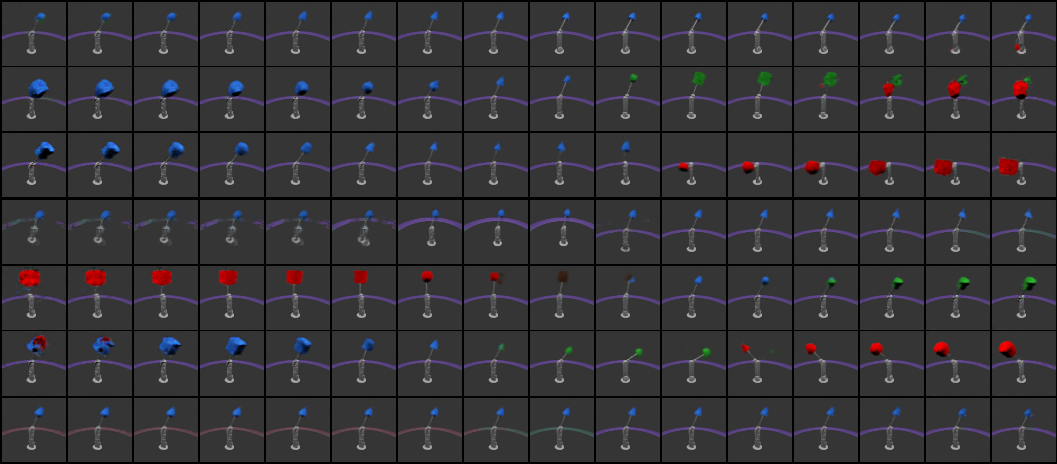}}
\end{figure}

\begin{figure}[!ht]
\floatconts
  {fig:beta-vae}
  {\caption{Traversed non-ignored latents of the trained \textbf{$\beta$-VAE} model  on a random sample of the \texttt{mpi3d\_toy} dataset.}}
  {\includegraphics[width=1.0\linewidth]{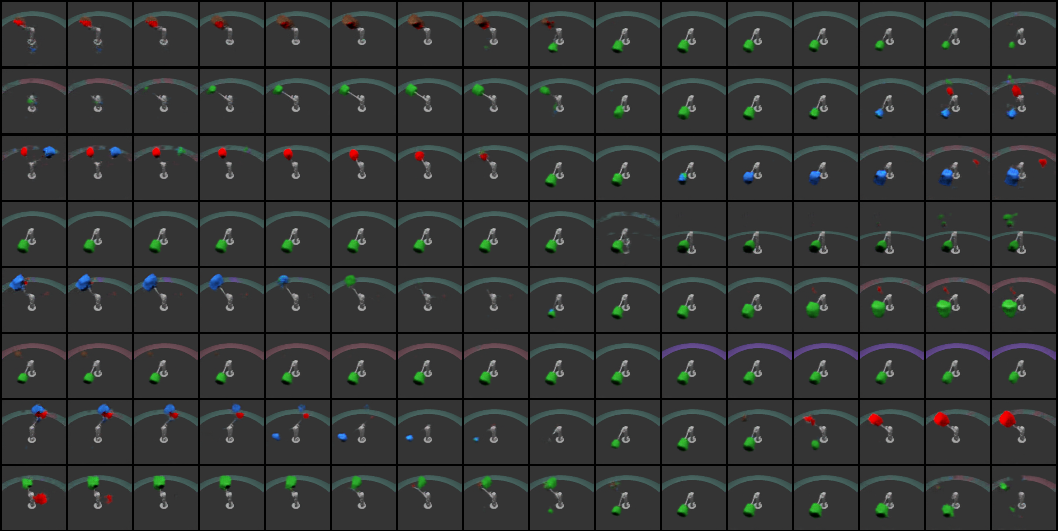}}
\end{figure}

\begin{figure}[!ht]
\floatconts
  {fig:factor-vae}
  {\caption{Traversed non-ignored latents of the trained \textbf{Factor-VAE} model  on a random sample of the \texttt{mpi3d\_toy} dataset.}}
  {\includegraphics[width=1.0\linewidth]{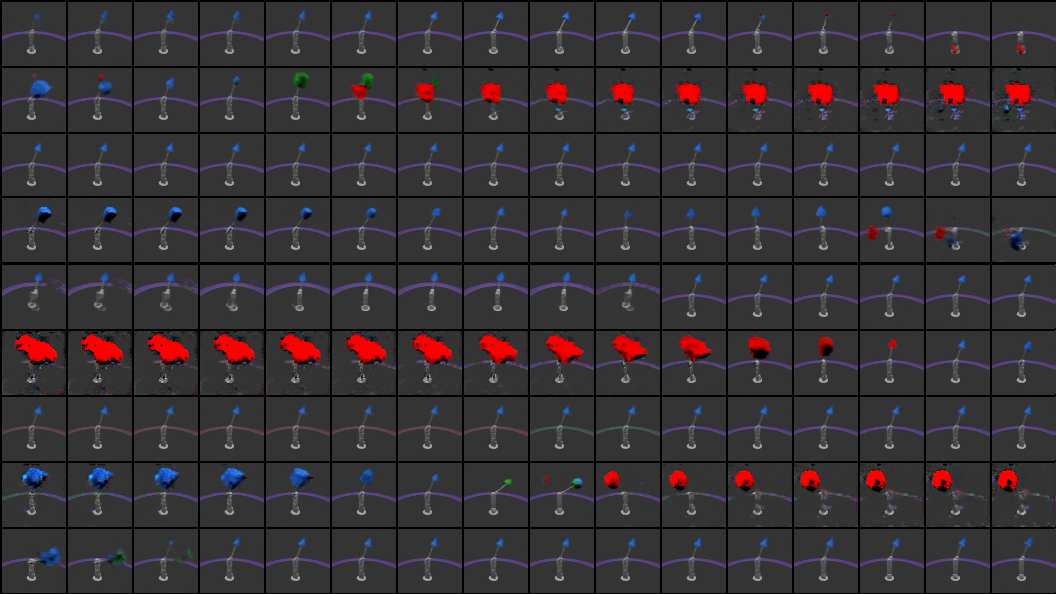}}
\end{figure}

\begin{figure}[!ht]
\floatconts
  {fig:dip-vae-i}
  {\caption{Traversed non-ignored latents of the trained \textbf{DIP-VAE-I} model  on a random sample of the \texttt{mpi3d\_toy} dataset.}}
  {\includegraphics[width=1.0\linewidth]{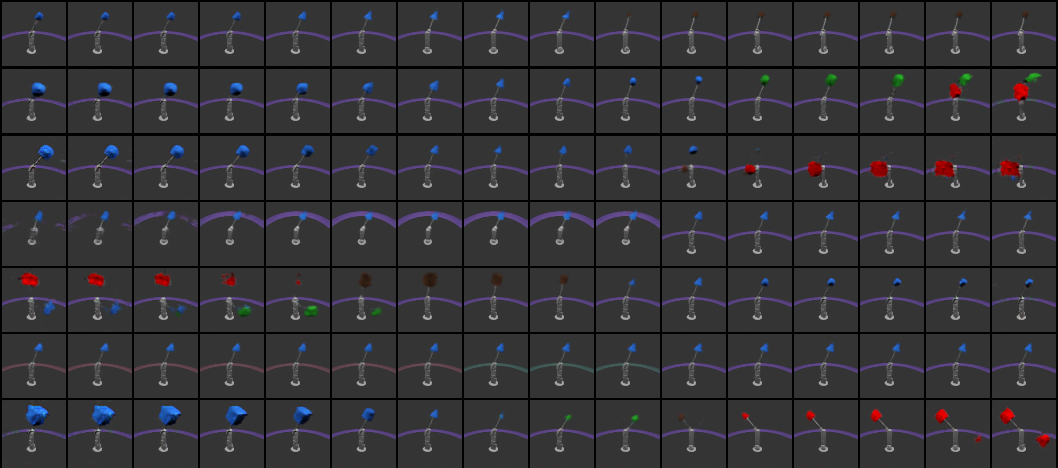}}
\end{figure}

\begin{figure}[!ht]
\floatconts
  {fig:dip-vae-ii}
  {\caption{Traversed non-ignored latents of the trained \textbf{DIP-VAE-II} model  on a random sample of the \texttt{mpi3d\_toy} dataset. This model surprisingly has twice as many non-ignored latent variables.}}
  {\includegraphics[width=1.0\linewidth]{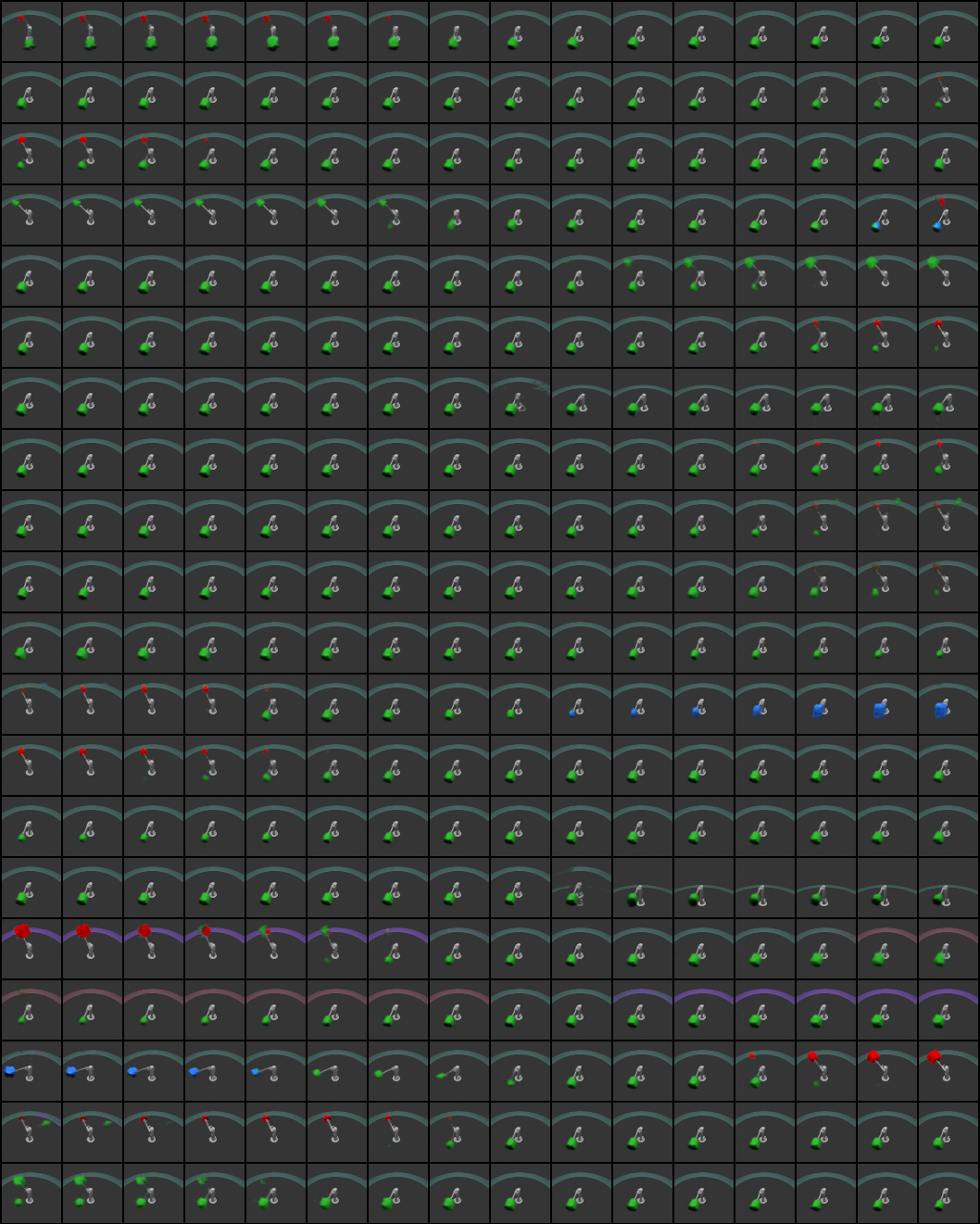}}
\end{figure}

\clearpage

\section{Progression of Evaluation Metrics During Training}\label{apd:training-log}

~

\begin{figure}[ht]
\floatconts
  {fig:subfigex1}
  {\caption{The progression of disentanglement evaluation metrics, \textbf{DCI, FactorVAE}, and \textbf{IRS},  throughout the training of the models.}}
  {%
  
    \subfigure[DCI]{\label{fig:DCI}%
      \includegraphics[width=0.65\linewidth]{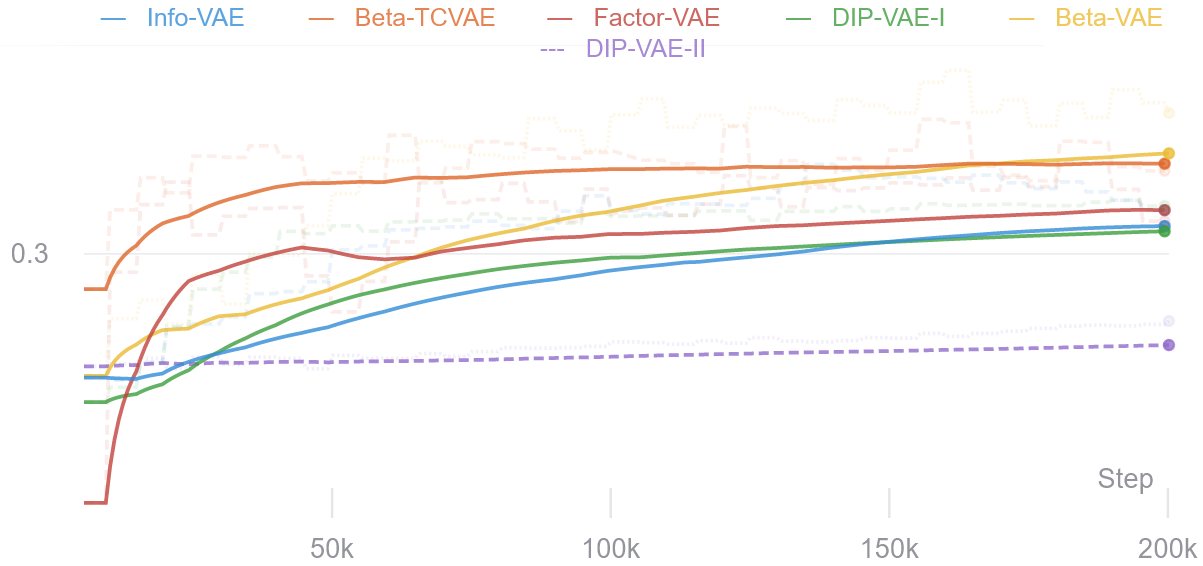}}%
  \vspace{10mm}
  
    \subfigure[FactorVAE Metric]{\label{fig:FactorVAE}%
      \includegraphics[width=0.65\linewidth]{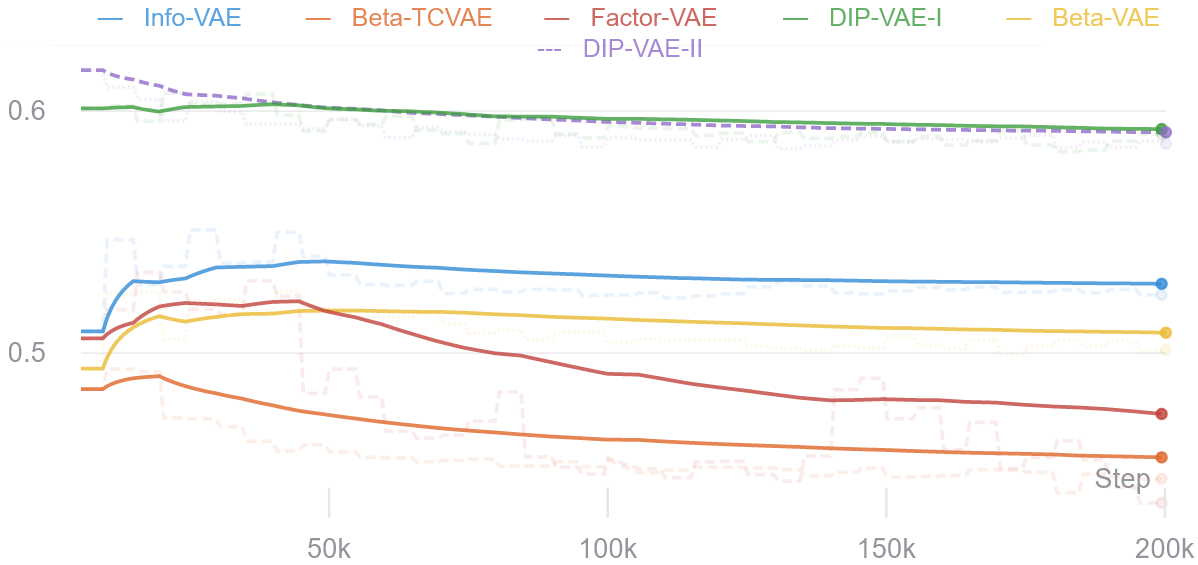}}
      \vspace{10mm}
      
    \subfigure[IRS]{\label{fig:IRS}%
      \includegraphics[width=0.65\linewidth]{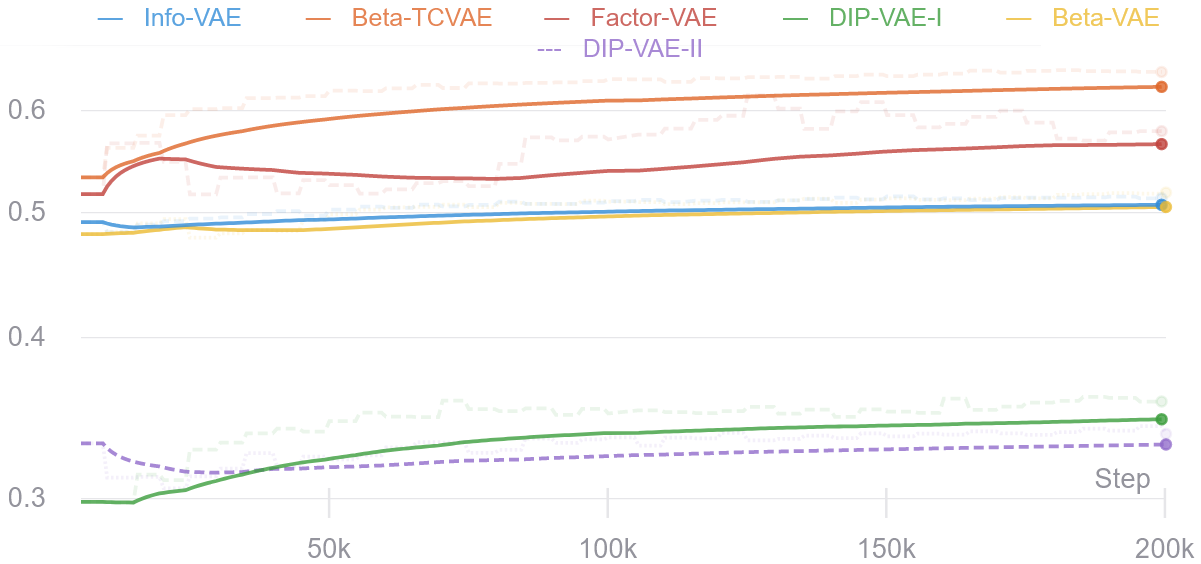}}
  }
\end{figure}

\begin{figure}[ht]
\floatconts
  {fig:subfigex2}
  {\caption{The progression of disentanglement evaluation metrics, \textbf{MIG} and \textbf{SAP},  throughout the training of the models.}}
  {%
    \subfigure[MIG]{\label{fig:MIG}%
      \includegraphics[width=0.65\linewidth]{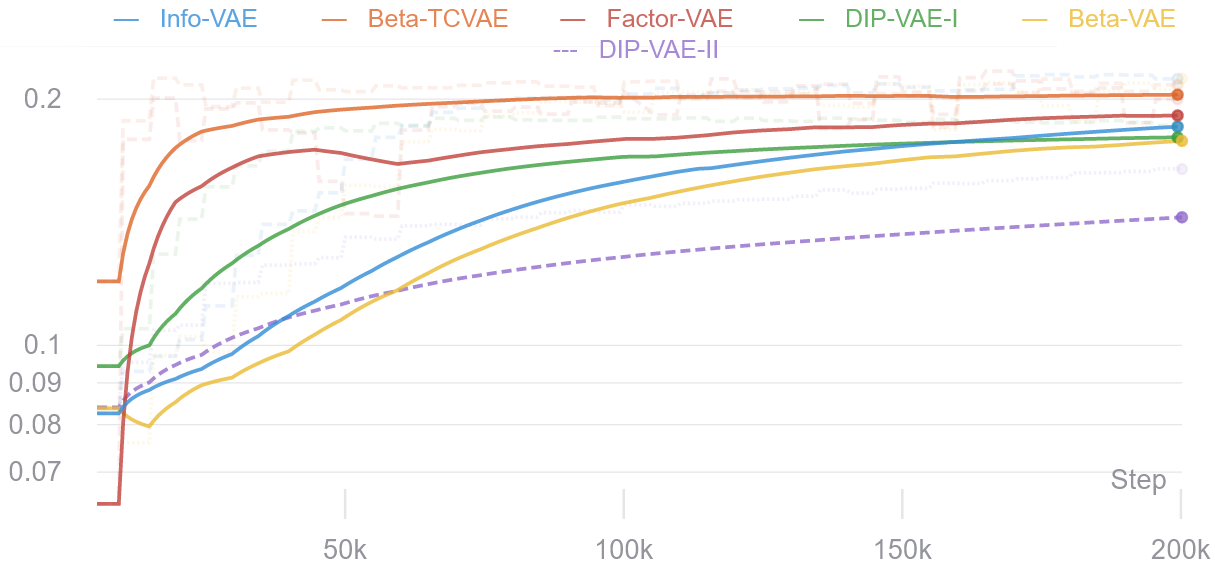}}
      \vspace{10mm}
    
    \subfigure[SAP Score]{\label{fig:Sap Score}%
      \includegraphics[width=0.65\linewidth]{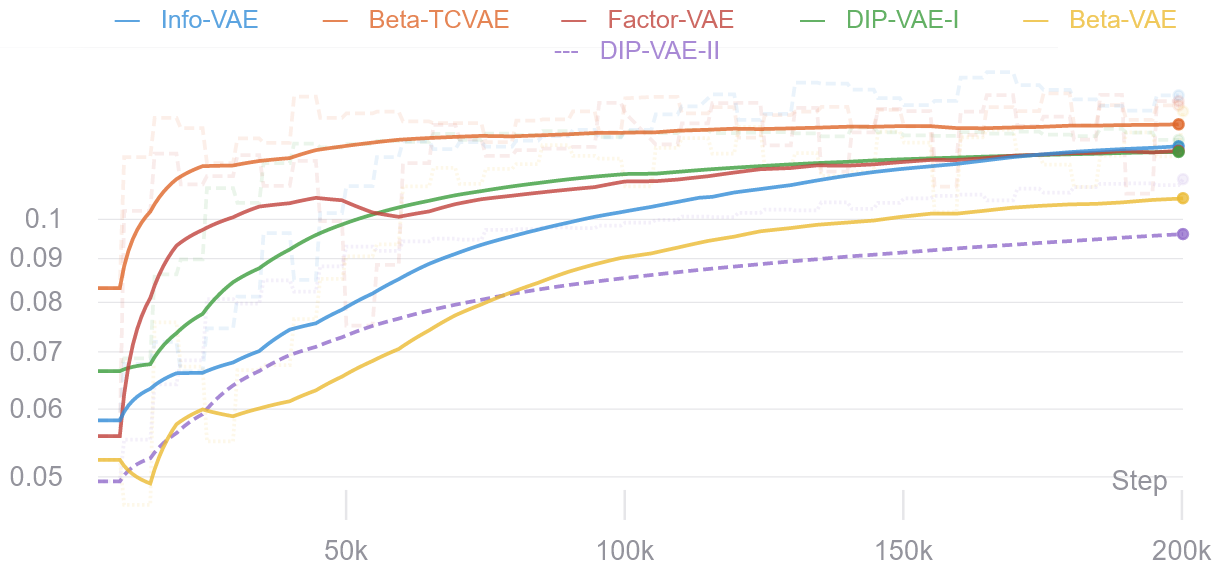}}
  }
\end{figure}

\end{document}